\documentclass[12pt]{article}
\usepackage{graphicx} 
\usepackage{geometry}
\usepackage{amsmath}
\usepackage[colorlinks=true,linkcolor=blue,citecolor=blue,urlcolor=blue]{hyperref} 
\usepackage{lineno}

\title{\textbf{Methods to improve run time of hydrologic models: opportunities and challenges in the machine learning era}}
\author {Supath Dhital*\\ \small{* The University of Alabama, Tuscaloosa, 35401, AL, USA}\\ \small{* sdhital@crimson.ua.edu}}
\date{}
\date{}

\newcommand{\keywords}[1]{\par\textbf{Keywords:} #1}

\begin{document}
\maketitle

\begin{abstract}
The application of Machine Learning (ML) to hydrologic modeling is fledgling. Its applicability to capture the dependencies on watersheds to forecast better within a short period is fascinating. One of the key reasons to adopt ML algorithms over physics-based models is its computational efficiency advantage and flexibility to work with various data sets. The diverse applications, particularly in emergency response and expanding over a large scale, demand the hydrological model in a short time and make researchers adopt data-driven modeling approaches unhesitatingly. In this work, in the era of ML and deep learning (DL), how it can help to improve the overall run time of physics-based model and potential constraints that should be addressed while modeling. This paper covers the opportunities and challenges of adopting ML for hydrological modeling and subsequently how it can help to improve the simulation time of physics-based models and future works that should be addressed. 
\end{abstract}


\section{Introduction}
Streamflow modeling and forecasting have been made using physics-based models in several cases. Since the pioneering advancement of the rational method in the middle of the $19^{th}$ century, hydrological models have gone through numerous stages: input-output models (black box), lumped conceptual models, and physically based distributed models \cite{mulvaney1851use}. Hydrologic models are used for plenty of uses, such as managing flood hazards, water resource planning, and ecosystem flow assessments. The study of hydrology is concerned with the space-time properties of the Earth's waters, including their occurrence, transport, distribution, circulation, storage, development, and management \cite{jain2017hydrological}. It has its cycle, with distinct procedures and meanings for each cycle. \\
\\
The hydrologic cycle is essentially the path taken by water as it travels through the land, the ocean, the atmosphere, and the earth in its several phases \cite{national1982scientific}. The cycle includes the movement, distribution, and storage of water on Earth. Multiple components make up the hydrological cycle: runoff (surface, interflow, and baseflow), precipitation, interception, evaporation, transpiration, infiltration, percolation, and moisture storage in an unsaturated zone \cite{jain2017hydrological}. In addition, human activity such as building dams, rerouting water, transferring basins, using groundwater or river water for agriculture, and monitoring runoff all affect the hydrological cycle \cite{chen2016population}. Hydrologic modeling is essential because of its significance and the increased understanding it provides for decision-making, especially in situations when there is a lack of data, insufficient understanding, or an inability to conduct prototype system experiments. Models may be quite useful in every component of the hydrological cycle, providing robust results more straightforwardly. The model can also be highly useful for determining the system reaction in anticipated/hypothetical circumstances. Big river basin hydrological simulations can be completed with physically distributed hydrological models, but their application is limited by the intricate nature of hydrological features. To manage water resources effectively in practice, however, easy-to-use, highly efficient hydrological models are required, and data-driven ML-based models have the potential to quickly map the relationships between meteorological predictors and hydrological responses without requiring in-depth descriptions of the corresponding physical processes \cite{yang2020physical}.\\
\\
In recent years machine learning outperforms on prediction capability and is loved because of its characteristics on simulation time. It overcomes the computational time burden of the hydrologic models. Without providing detailed explanations of the underlying physical process, it may rapidly anticipate future possibilities and give mapping connections between predictors and hydrological reactions by identifying previous trends \cite{adnan2019daily, beven2020deep, kasiviswanathan2016potential}. Studies have demonstrated in some cases the ML models can outperform other traditional process-based techniques \cite{kratzert2019benchmarking}. Quantitative correlations between a response variable and one or more explanatory factors are frequently described using regression analysis. With a hydrological focus, this model is helpful in identifying relationships between precipitation and runoff for both the same and various watersheds \cite{xu2017black}. A Deep Learning (DL) model may be able to capture the behavior if there are consistent anomalies between the conceptual framework of a hydrological model in a certain catchment and the characteristics of the hydrological processes in that catchment \cite{beven2020deep}. Researchers also tried physics-informed machine learning to capture physical behaviors and integrated it into the ML models which can surpass the limitations of singular models, which yields more robust outcomes \cite{xu2022hybrid, ren2023enhancing}. \cite{yang2023monthly} carried out in the glacial basins have shown that hybrid models—which combine standard hydrological models (GSWAT+) with DL-based models—have major advantages. While data-driven methodologies yield promising findings in a short simulation time, there are implementation-related issues that must be addressed to get optimal outcomes. This paper provides a brief description of the opportunities and problems associated with machine-learning techniques that use to improve the computing efficiency of hydrologic forecasts.

\section{Challenges in runtime efficiency with physics-based hydrologic models}
Lumped, semi-distributed, and distributed models are the three categories into which hydrological models can be divided based on how the basin is represented spatially \cite{yoosefdoost2022hydrological}. Distributed models use a grid or mesh to depict the watershed, semi-distributed models split it up into a small number of sub-basins or hydrological response units, and lumped models show the watershed as a single unit. The purpose for which a hydrologic model is constructed dictates its design and structure and the architecture of model reflects the uncertainties. \\
\\
A physically based distributed catchment model has rather high computational and data needs to run. Because computations are made for several grid points at each time step, these include a big storage capacity and fast processing speed which limits the necessities of rapid modeling \cite{teng2019enhancing}. Moreover, The parameters must be calibrated using observed data or another means to get reliable findings. To minimize the error between these two sequences, it is accomplished by repeatedly changing the model's parameters and assessing one or more objective functions that measure the fitness between the simulated and observed variables. The effectiveness of the optimization process depends on an effective parameter space exploration, but for distributed models, this may be quite computationally expensive due to the amount of information in the spatial representation and the requirement to solve complicated partial differential equations \cite{bermudez2019rapid}. So optimized model calibration and configuration is needed \cite{mudashiru2021flood}. Furthermore, an empirical model is not suitable for all catchments since different hydrological processes are not taken into separate account in these models, it is difficult to permit changes in the relative influence of those processes. Rather than using complex and tedious Numerical Weather Prediction (NWP), the Machine Learning approach will be the best alternative for the short computational time with efficient results \cite{dhital2024forecasting}.\\
\\
Hence it need to be resolved and ML can potentially improve some sorts of challenges. The continuous increase of the computational power allowed more complex ML networks in a variety of hydrological applications, especially in situations where the classical approach is computationally demanding such as for river flow prediction \cite{xu2020using}, water resources management \cite{rozos2019machine}, flood inundation mapping \cite{lin2020prediction} etc.

\section{Opportunities with ML}
Machine learning offers as a powerful toolkit for tackling complex problems in various domains, including hydrology. Either by surrogate machine learning modeling \cite{fraehr2022upskilling} or by offering an adaptive sample the input data focusing on areas or times where the model's predictions are most uncertain, ML reduces the overall computational cost with the same result.\\
\\
Running data-driven black box models requires less computational power since most of the models do not require a lot of processing resources. Artificial Neural Networks (ANNs) can generalize the structure deep within the whole dataset and derive the relationship between a process's inputs and outputs without explicitly providing them with the laws of physics \cite{xu2017black}. The capability of ANNs to infer solutions from data without the need for previous knowledge of the regularities in the data adjusts solution over time to account for changing conditions, parallelization computation can be done \cite{zealand1999short}. In addition, in time series applications, Long Short Term Memory (LSTM) works incredibly well. It also has the benefit of allowing each LSTM unit to be assigned a dynamic state, which acts as a memory mechanism to speed up and parallelize computations. For streamflow simulations, an LSTM-based model outperformed a parsimonious approach in the field of hydrologic modeling in terms of generalization capability \cite{ayzel2021development}. \cite{kratzert2019towards} proposed an Entity-Aware-LSTM model, which was designed with the ability to include the features of the catchment. It performed better than both the regionally and basin-specific hydrological models. Unlike traditional basin-specific models that require calibration for each individual catchment, this model offers learning from catchment features and makes it adaptable to unseen basins, potentially offering good performance without the need for calibration for new locations. One of the key problems in hydrology, simulating accurate runoff \cite{kratzert2018rainfall} however, state of an art model developed by \cite{fan2020comparison} is able to accurately predict runoff. So, ML can potentially do a lot of task that are complicated for physics-based models because of its flexibility to play around with various datasets. Furthermore, \cite{wu2024improving} created a coupled hydrological model (WetSpa-LSTM) by fusing a deep learning model (LSTM) with a distributed hydrological model (WetSpa-WW). The coupled model outperformed the conventional hydrological model, according to its results. In \cite{bindas2024improving}, a novel differential routing technique was developed to infer parameterizations for Manning's roughness and channel geometries from raw reach scale attributes like catchment areas and sinuosity. This technique mimics the classical Muskingum-Cunge routing model over a river network by integrating it with neural networks (NNs).\\
\\
Moreover, ML offers optimization of the model by calibrating the parameters too. Several techniques may be used for calibration optimization, which is necessary to get good results. The two strategies that have historically been used are local search techniques and global search techniques \cite{yang2019methods}. While global search methods, such as evolutionary algorithms, use stochastic search mechanisms to overcome the limitation, the number of model runs increases with the number of parameters, which results in a slow convergence rate and high computational cost \cite{zhao2015artificial}. Local search algorithms, such as gradient descent, are designed to identify local minima within a parameter space, but cannot find global optimum \cite{kavetski2018parameter}. The most effective option is to use meta-modeling or surrogate modeling techniques, which concentrate on substituting a more straightforward and economical surrogate model developed from statistical or data-driven procedures for the original numerical model. Least squares support vector machines (LS-SVMs) \cite{bermudez2019rapid} and artificial neural networks (ANNs) are two common surrogate models \cite{gu2020surrogate, chu2020ann}.
 \\
\\
By inferring all aforementioned progress in the field of hydrology either completely replicated by a data-driven approach or coupled with the physics-based hydrologic model, machine learning provides parallelization computation, simplicity of a data-driven approach, and model optimization and calibration capability. While talking about the advantages, it does contain some pitfalls and has a couple of challenges to implement and model techniques. 

\section{Challenges in implementing ML}
In the fascinating era of ML, researchers are busy with proposing new models and approaches to make robust outcomes possible. While applying it for various applications, it does contain some implementation challenges related to data availability and quality, resampling various techniques, inconsistent validation procedures, model interpretability, and transparency, generalization issues, computing resource requirements for complex model architecture, and so on.\\
\\
Hydrologic modeling and forecasting is a rapidly evolving area, with one of the fastest-growing concerns being machine learning methods. However, there are several difficulties in developing and evaluating the effectiveness of data-driven machine-learning models for these kinds of applications. While encompassing all the differences related to basin physical behaviors and modeling, machine learning (ML) can perform better in hydrological applications, but only if the cost is higher computational complexity \cite{rozos2019machine} as it needs various input parameters from various sources and in the case of DL, multiple layers and neurons makes architecture more complex. Even though the model works well, \cite{yang4720569combining} notes that the interpretable hybrid model improves the validity and precision of water balance assessment by substantially improving the accuracy of runoff residual predictions and providing logical feature explanations. However, the transformer (TF) model's complex architecture leads to increased computational demands. ML model so-called data-driven model completely relies on the input data so little blunder in the input itself lead to huge error results in overfit or underfit. \\
\\
There are various limitations to ML-based hydrological models. One of them it, that to get robust performance, a substantial quantity of training data is needed \cite{kratzert2019benchmarking}. Secondly, because long-term hydrological measurements are lacking in most streams worldwide, the usefulness of ML in hydrological simulations and forecasts is limited. Researchers even attempted to utilize transfer learning to model the ungauged basin after modeling one gauged basin with great accuracy \cite{jiang2018computer}. However, the model's overall performance was not accurate in extreme events or diverse terrain setups. Furthermore, ML models frequently use data aggregated throughout the watershed as model inputs, ignoring the spatial heterogeneity in the watershed landscape and the associated atmospheric forcing \cite{jiang2018computer}. Nonetheless, hydrological reactions are frequently greatly influenced. Since the frequency of high flows in streamflow time series is relatively low, incorporating high flows into training data is not always feasible and results in inadequate model training. ML models are solely based on observational data, which leads to poor simulations during high flows \cite{yang2019real} subsequently they cannot perform well in extreme events. Not only problems in data inconsistency and input parameters, but challenges are there in physics-informed machine learning too.\\
\\
The primary uses of ML or DL in hydrology include learning and extracting intricate data distributions and modeling hydrological processes using observable data which scrap hydrological information from images \cite{shen2018transdisciplinary}. But it was often criticized for having features that made them seem like black boxes. While there are some insights gained from comparing the LSTM cell states and sliding windows employed in the LSTM model architecture with the corresponding meteorological parameter as the dynamic catchment features and not possible for all modeling approaches, this is laborious work whose rationality has not yet been shown. Choosing the optimal deep learning model to build can also be more of an art than a science because different model architectures might disclose different discoveries from the same volume of data. This makes deep learning challenging at times. Using delayed parameters to capture patterns of any meteorological parameters, makes the model more complex, and manually doing such action for a large dataset and multiple basins will be tedious and time-consuming.

\section{Methods to improve runtime efficiency with ML}
Accurately and timely hydrological simulations are vital for water resource management, forecasting, and so on. Fully distributed process-based hydrologic models, while powerful, often require significant computational resources due to their intricate equations \cite{teng2019enhancing}. This poses a challenge to the rapid assessment of watersheds, however, the emergence of machine learning offers a transformative approach to improve the runtime efficiency of hydrologic modeling.\\
\\
Hydrologic models often deal with high-dimensional datasets which include meteorological-driven as well as topographical factors. Though it offers a detailed picture of the hydrologic system, it also presents a computational challenge. Dimensionality reduction using principal component analysis (PCA) is one of the effective methods that reduce the volume of a dataset by creating new covariates that aren't related to each other without losing essential information \cite{hasan2021review} that is crucial for modeling and performance. By extracting key spatial features for spatial dimension datasets with Empirical Orthogonal Function (EOF), which have been used in the field of spatial data, especially in remote sensing and climate change \cite{alvarez2016predicting, ghosh2021rationalization}. Feeding ML models to lower dimensional data automatically leads to faster convergence toward findings. Moreover, the robustness of the findings and computing time will be ensured if the model structure is sufficiently homogeneous and simple inside the model-building framework \cite{quilty2020stochastic}. Using high-resolution grids to enhance computationally efficient hydrologic model results using ML also poses some great potential to enhance the simulation time as well as accuracy. To simulate monthly streamflow predictions, \cite{xu2024coupling} combined a physically based hydrological model with deep learning. First, a simplified hydrological model is created by selecting grid cells from a distributed hydrological model optimally based on soil moisture characteristics. using a deep learning model that predicts future monthly forecasts by combining the model-simulated predictions with additional predictors from other sources. This intentional hybrid model saves decision-makers time and effort by significantly reducing the computing overhead of rolling predictions when compared to dispersed hydrological models.\\
\\
Another useful technique is parallel computing with machine learning. It provides several notable benefits, including the ability to run simulations more quickly and enhance model calibration for completely distributed models. These days, the effort is distributed across numerous processors through the use of parallel computing and machine learning-based hydrologic models, greatly reducing the simulation duration. Parallel computing aids in the calibration of hydrologic models by enabling the rapid exploration of this parameter space through the execution of several simulations with various parameter combinations, ultimately leading to the best possible calibration procedure. According to \cite{asgari2022review}, the efficiency of hydrologic models may be accelerated up to a certain point by the number of parallel processing units. Using parallel computing, \cite{yen2014c} was able to improve run time for the Soil and Water Assessment Tool (SWAT) by 28–35\%. Similarly, \cite{seong2015automatic} was able to speed up the spatially-lumped hydrologic model Hydrologic Simulation Program FORTRAN (HSPF) by 47–67\% with two and four processors, respectively. Currently, using parallel computing to train machine learning, or more specifically, focusing on the iterative process of deep learning, would significantly improve the performance of deep learning as it processes data and updates model parameters millions of times. Optimizing machine learning performance also requires tuning hyperparameters; parallel computing makes it possible to run numerous hyperparameter configurations concurrently, which accelerates the search for the optimum configuration. By employing a swarm intelligence-based artificial neural network to anticipate hydrological time series many steps ahead, \cite{niu2021parallel} improved the Nash-Sutcliffe efficiency value by 6.4\%. Therefore, it may also enhance the ML model's predicting performance in addition to run time. \\
\\
In addition, adopting feature engineering helps to sort the importance of parameters in the model so, removing the least significant parameters significantly improves the performance as it directly helps to reduce the number of trainable parameters for the ML and DL. 

\section{Conclusion}
Accurate runoff forecasting techniques are becoming more and more necessary for water resources due to the severity of climate change in recent years. The integration of machine learning into hydrological modeling offers a transformative approach to tackling several challenges of process-based hydrological models. By leveraging techniques like dimensionality reduction of the dataset, parallel computing, adaptive sampling, and feature engineering, ML paves the way for significantly faster and more efficient hydrologic simulations which will not only save time but also provide robust results during the necessities of the rapid assessment. \\
\\
However, it also has some sort of implementation pitfalls and implications that should be addressed to get the optimum solution. Need for a diverse range of input datasets- which is difficult to get for some study areas as the lower frequency of historical data, tuning hyperparameters while adopting transfer learning, complex model architecture to get accurate results, and lack of rationality of how black box model works are the main challenges on applying ML while modeling hydrology. To ensure that the data used in the ML model validation is different from that used in the model training, a variety of cross-validation techniques can be employed. In many cases, modeling machine learning on networks with quite complicated topography necessitates a large amount of processing power and CPU time during training. Despite these challenges, the synergy between hydrology and machine learning holds immense scope and impact in the future. 

\section{Future works}
The significant improvement in the fusion of ML and hydrology is flourishing and demonstrated improved runtime efficiency of hydrologic models, exciting opportunities for further exploration remain. Some recommendation can be listed as follows. 
\begin{enumerate}
    \item Precise evaluation of the water balance is essential for managing water resources sustainably and reducing the negative impacts of climate change and human activities on local hydrology. 
    \item Concentrate on the interpretable hybrid model's correctness and performance to improve the ML models' overall run time. 
    \item The hybrid model's flexible model coupling technique guarantees its application in several sectors, suggesting wider regional applications that require further verification. Although the present studies span an extensive region, it is difficult to apply these results to bigger basins or at regional scales, which limits their usefulness as a border reference. Therefore, in order to assess the scalability of the suggested adjustments, future studies should focus on expanding the scope of balance correction analysis to encompass bigger areas. 
    \item To systematically evaluate a suite of parallelization scenarios for improving speedup gain, efficiency, and solution quality, different combinations of hydrologic models, optimization algorithms, parallelization strategies, parallelization architectures, and communication modes must be implemented \cite{asgari2022review}. If this can be accomplished using various machine-learning techniques, it would have a significant impact on the field of hydrology. Speedup gain/efficiency decreases as the number of parallel processing units increases, particularly after a threshold.
    \item In order to apply and assess the usefulness of data-driven approaches under various climate and geomorphological conditions, comparisons between process-based hydrologic models and machine learning have been made in limited climate and terrain characteristics. These comparisons have been made by adding more complex and variable conditions, such as snow-dominated watersheds, mountainous terrain, river routing modeling in dam-placed conditions, and so on.
    \item In the context of ensemble hydrologic models and physics-informed machine learning models, in particular, parallel computing appears to be not applied. Thus, research on that subject significantly lessens significant barriers to ML use in hydrology. It can automate laborious tasks like hyper-parameter tuning in multiple threads at once, assist in sticking to a small dataset so that, after applying parallel computing, it can even work with high-dimensional data without losing dimension, expedite training and validation procedures, and assist in simultaneously determining the best model architecture and performance. 

\section{Acknowledgement}
I would like to thank Dr. Mukesh Kumar and Dr. Sagy Cohen for the support and guidance. 

\end{enumerate}
\bibliography{reference}
\end{document}